\documentclass[nohyperref]{article}

\usepackage{microtype}
\usepackage{graphicx}
\usepackage{subfigure}
\usepackage{booktabs} % for professional tables
\usepackage{xcolor}

\usepackage{hyperref}

\usepackage[accepted]{icml2022}

\usepackage{amsmath}
\usepackage{amssymb}
\usepackage{mathtools}
\usepackage{amsthm}
\usepackage{lipsum}
\usepackage{balance}
\usepackage{enumitem}

\usepackage[T1]{fontenc}

\usepackage[capitalize,noabbrev]{cleveref}

\theoremstyle{plain}

\theoremstyle{definition}

\theoremstyle{remark}

\usepackage[textsize=tiny]{todonotes}
\icmltitlerunning{Interpretability by design using computer vision for behavioral sensing in child and adolescent psychiatry}

\begin{document}

\twocolumn[
\icmltitle{Interpretability by design using computer vision for behavioral sensing in child and adolescent psychiatry}

%alternative
%Automation of coding child and adolescent behavior
%in psychiatry



\icmlsetsymbol{equal}{*}

\begin{icmlauthorlist}
\icmlauthor{Flavia D. Frumosu}{equal,dtu}
\icmlauthor{Nicole N. Lønfeldt}{equal,CUH}
\icmlauthor{A.-R. Cecilie Mora-Jensen}{CUH}
\icmlauthor{Sneha Das}{dtu}
\icmlauthor{Nicklas Leander Lund}{dtu}
\icmlauthor{A. Katrine Pagsberg}{CUH,sch,gcb}
\icmlauthor{Line K. H. Clemmensen}{dtu}
\end{icmlauthorlist}

\icmlaffiliation{dtu}{Department of Applied Mathematics and Computer Science, Technical University of Denmark, Kongens Lyngby, Denmark}
\icmlaffiliation{CUH}{Child and Adolescent Mental Health Center, Copenhagen University Hospital – Mental Health Services CPH, Hellerup, Denmark}
\icmlaffiliation{sch}{Department of Clinical Medicine, Faculty of Health and Medical Sciences, University of Copenhagen, Copenhagen, Denmark}
\icmlaffiliation{gcb}{Department of Clinical Biochemistry, Hospital Glostrup - University Hospital, Glostrup, Denmark}

%\icmlcorrespondingauthor{Line H. Clemmensen}{lkhc@dtu.dk}
\icmlcorrespondingauthor{Flavia D. Frumosu}{fdal@dtu.dk}

\icmlkeywords{Machine Learning, ICML}

\vskip 0.3in
]



\printAffiliationsAndNotice{\icmlEqualContribution} % otherwise use the standard text.

\begin{abstract}
Observation is an essential tool for understanding and studying human behavior and mental states. However, coding human behavior is a time-consuming, expensive task, in which reliability can be difficult to achieve and bias is a risk. Machine learning (ML) methods offer ways to improve reliability, decrease cost, and scale up behavioral coding for application in clinical and research settings.
Here, we use computer vision to derive behavioral codes or concepts of a gold standard behavioral rating system, offering familiar interpretation for mental health professionals. Features were extracted from videos of clinical diagnostic interviews of children and adolescents with and without obsessive-compulsive disorder. Our computationally-derived ratings were comparable to human expert ratings for negative emotions, activity-level/arousal and anxiety. For the attention and positive affect concepts, our ML ratings performed reasonably. However, results for gaze and vocalization indicate a need for improved data quality or additional data modalities.
\end{abstract}

\section{Introduction}
Computer vision has the potential to aid mental health professionals establish diagnoses and monitor progress of treatment. Visual observations are an important clinical tool as many psychiatric diagnoses are characterized by either increased or decreased motor activity (e.g., attention deficit
hyperactivity disorder (ADHD), anxiety disorders, or depression \cite{mendes2018fine,american2013diagnostic}.
% (e.g., attention deficit hyperactivity disorder, mania, depression), distinct gestures such as hand flapping (autism spectrum disorders (ASD)), avoidance of eye contact (ASD) and decreased facial expressiveness (schizophrenia and ASD).\\ 
However, not all psychiatric disorders have such distinguishing signs. For example, obsessive compulsive disorder (OCD) is characterized by intrusive, repetitive thoughts or actions. The internal processes cannot be directly observed and may be especially difficult for children to describe \cite{thapar2017rutter}.
% Compulsions, repetitive or ritualized behaviors that the individual feels compelled to perform (e.g. hand washing, checking, turning light switches on and off), can be directly observed, but they are not always performed in the clinic. 
Thus, monitoring more general expressions of emotions and other mental states and processes holds important clinical information. 
%\textcolor{red}{I do not think the following sentence fits well. It would also require a more explaining of how what we are doing in this study is connected to prevention and early treatment.}An early treatment can help in preventing the development to a chronic stage. \cite{waltereit2018adolescent}. 
For example, facial expressions provide important information about moods, emotions and cognitive effort \cite{barrett2016handbook}; eye contact or gaze can provide information about how engaged a person is and the quality of rapport between clinician and patient \cite{montague2011modeling}. 
% Thus, psychiatry would also benefit from tracking more general facial and body actions. 
However, systematically recording behavioral observations is a labor-intensive process for humans. Machine learning (ML) methods have the possibility to automate this process resulting in decreased labor and increased efficiency in psychiatric and behavioral research settings.

% %\subsection{Related Work}
% Behavioral measures from videos recorded of free-arrangement, arrangement on contrasting carpets, and hand washing tasks have significantly correlated to CY-BOCS scores of youth with OCD and healthy controls \cite{bernstetin2017}, indicating that measurement of behavior using video recordings, at least at a task related level, is a valid approach for quantifying OCD psychopathology. To the best of our knowledge, there are no investigations of the use of video for quantifying OCD psychopathology at a finer level, e.g., using gestures and facial expressions. In the following, we review existing tools which provide behavioral measures at this finer granular level. \\
% \textcolor{red}{ Need to mention, AUs - FACS, gaze OpenFace, activity, motion in a video optical flow, person activity}
Current state-of-the-art computing tools for emotional expression with video analysis include convolutional neural networks with a focus on the prediction of action units with the Facial Action Coding System (FACS) \cite{grabowski2019emotional,jiang2022utilizing,washington2020data,de2020computer}. Pre-trained models, like OpenFace \cite{Baltrusaitis2018}, are primarily trained on adult as opposed to child expressions \cite{abbasi2022}. Previous studies have used eye tracking systems to link eye movements and attention from children with ASD \cite{liu2016, de2020computer}. However, as we are analyzing historical data, no eye tracking devices have been employed, which comes with different challenges \cite{Cristina2018}. 
% Pervasive gaze tracking (tracking \emph{in the wild}) comes with several challenges, which also hold for our study: low-resolution images, limited training data requiring person-independent gaze estimation, and reduced calibration \cite{Cristina2018}. 
% In our case, no calibration was performed, as historical data are used.
Within human motion tracking, several methods focus on background subtraction and optical flow \cite{Godbehere2012,kajabad2019people,lee2017robust}. We used a background subtraction approach in our work due to its simplicity. 

In this work, we aimed to design an interpretable ML approach that can give actionable feedback to clinicians by learning individual, interpretable, concepts. Inspired by recent research with concept bottleneck architectures \cite{Koh2020} and prototype concepts \cite{Chen2019}, we show how interpretability can be achieved by design. Our design draws direct connections between gold standard, individual codes of human behavior and behavior ratings derived from computer vision models. We test the performance of ML-derived codes by estimating their agreement with human ratings of youth behavior and our transparent design allows for evaluation by concept.

\section{Methods}
This work is a part of a larger study. A detailed description of our methods and analysis plan are outlined in a statistical analysis plan \cite{lonfeldt2022computational}. 
\subsection{Video data set}
Children and adolescents (8-17 years) with (n=25) and without OCD (n=12), who participated in a case-control study and randomized clinical trial, completed a diagnostic interview, the Kiddie Schedule for Affective Disorders and Schizophrenia (K-SADS) \cite{puig1986kiddie}, before inclusion in the study. The K-SADS is a semi-structured interview used for early diagnosis of psychiatric disorders in youth between the ages of 6 and 18 \cite{birmaher2009schedule}. Participants in this study did not have clinical nor subclinical tics nor hyperactivity. We sampled 30 seconds from the depression and mania portions of the interviews resulting in 74 videos. The videos have not been pre-processed. The diagnostic interview was video recorded using a Sony video camera in the mental health center for patients and a research unit for controls. The camera was not placed in a uniform position in the room  across participants' videos. All cameras were focused on the youth as opposed to the interviewer, who only sometimes appeared in the shot.  
\subsection{Behavior concepts: Coding Interactive Behavior (CIB) adolescent version}
The youth behavior in the videos was coded using the adolescent version of the Coding Interactive Behavior (CIB) manual \cite{feld1998cib}. The CIB is a global rating system, in which items are scored from 1 to 5 and half-points can also be assigned (e.g., 2.5 or 4.5). Higher scores indicate higher frequency, duration and intensity of a behavior. 
Studies have demonstrated that CIB scores of children with and without psychiatric diagnoses differ significantly \cite{feldman2012cib}.\\
We chose 7 items from the CIB manual with focus on youth behaviour and the relationship between the youth and the interviewer as the concepts for our design. The items evaluate youth engagement and emotional states, which are observable in 30-second intervals. The items are inherently valuable, however they can also contribute to assessing youth distress, therapeutic alliance and parent-child synchrony.

The chosen concepts with the corresponding definitions of high scores are presented below: 
\begin{itemize}[noitemsep,topsep=0pt,parsep=0pt,partopsep=0pt]
    \item \textbf{Gaze}: The child consistently looks at the interviewer.
    \item \textbf{Vocalization}: The child speaks frequently, for a long duration and can express themselves well.    
    \item \textbf{Positive affect}: Signs include smiling, laughing, calmness and seeming interested.
    \item \textbf{Negative emotionality}: Signs include expressions of anger, sadness (yelling, cursing, crying).
    \item \textbf{Activity-level/arousal}: Talking quickly, loudly or with vocal fluctuations, or high levels of body movement and facial expressiveness.
    \item \textbf{Anxiety}: Explicit signs of nervousness i.e., darting eyes, inexplicable enthusiasm, long silences, fidgeting, sudden changes in emotion, anxious statements.
    \item \textbf{Attention}: The child is focused on the interview, cooperates and gives relevant answers.
\end{itemize}

\subsection{Behavior rating methods}
\subsubsection{Human raters}
Two mental health professionals (Raters 1 and 2), co-authors, trained in using all the codes in the CIB on 3-minute videos reached 89\% percent agreement on a separate set of videos. 
Raters 1 and 2 scored the behavior of the youth in the clinical interview video samples using the 7 CIB items previously presented. Raters used vision and audio to assign scores. Due to the time-consuming process of the scoring, each rater scored 44 videos in a random order. From these 44 videos, 14 videos were scored separately by both raters. All the 74 videos were scored by either Rater 1 or 2. Raters scored a batch of 7 videos at a time (the last session had 9 videos). The raters met and discussed codes after each batch to avoid coder drift. It was attempted to blind raters to diagnostic status and diagnostic interview chapter, but this information is often indirectly available in the video.

\begin{figure*}[h!]
\centering
\includegraphics[width=12cm]{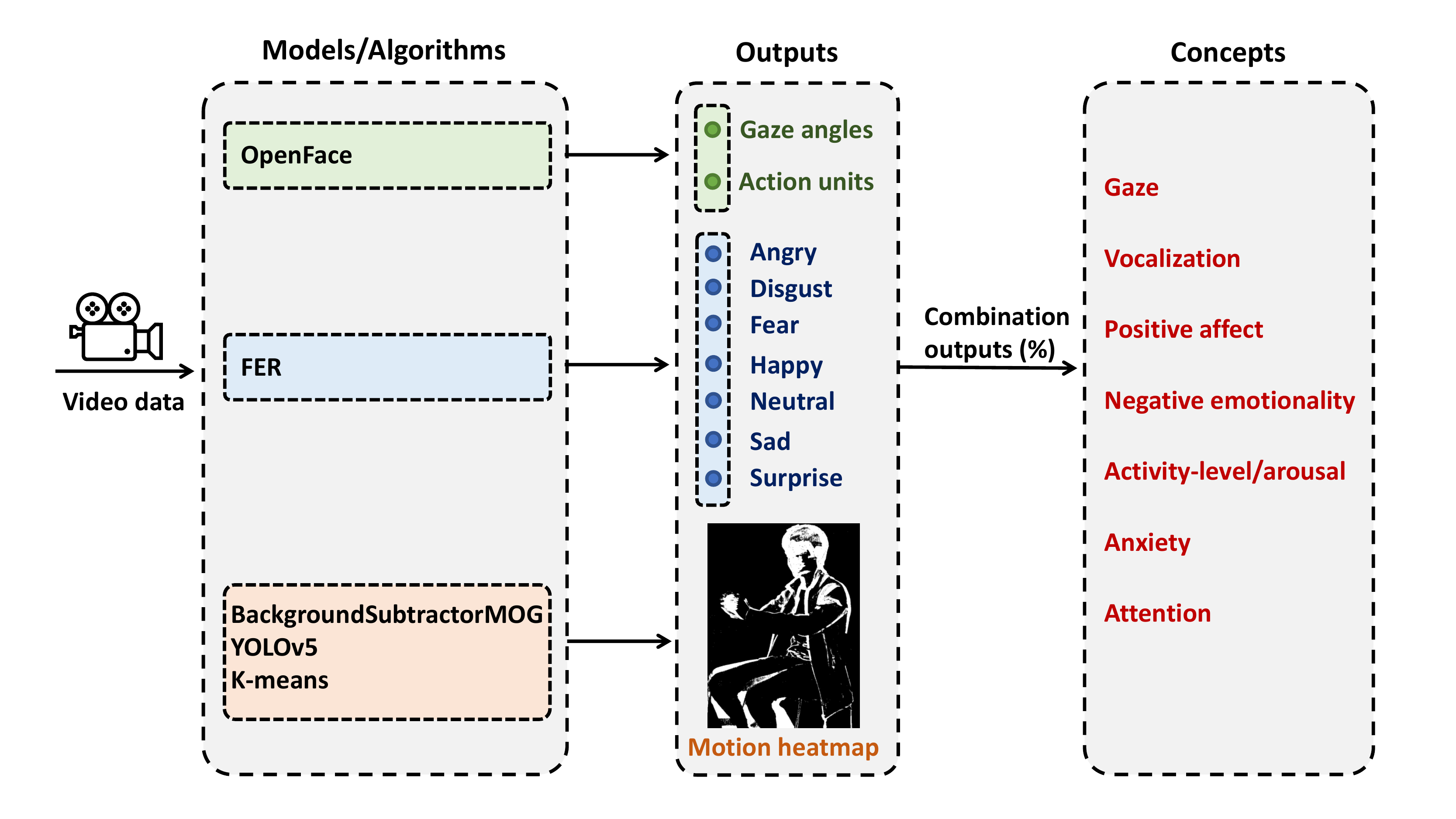}
\vspace*{-3mm}
\caption{Methodology: Each video is fed into the models/algorithms which return different outputs. These outputs are further transformed into percentages and are combined to describe the concepts.}
\label{fig:methodology}
\end{figure*}

\subsubsection{Machine learning methods}

Through the use of behavior codes as concepts, we show that interpretability can be achieved by design, Figure \ref{fig:methodology}. Face landmark detection and facial expression recognition pre-trained models were used to extract features or to predict youth expressions from the videos. It is worth mentioning that the models have been trained on adults - not on children or adolescents. Time-wise, the models are slow for longer videos, however, pre-processing of the videos can be performed in order to obtain the outputs faster.\\
For activity quantification, we detected the entire body posture and we created a motion heatmap based on the accumulation of changed pixels over time.\\
Composite scores were created using the outputs from the pre-trained models to match the concepts. The matching was done within our interdisciplinary team including mental health professionals.
\begin{table*}[h!]
\centering
\caption{For reproducibility purposes the used models/algorithms are presented along with the used hyperparameters.}
\label{table:hyperparameters}
\resizebox{2\columnwidth}{!}{
\begin{tabular}{llll}
\hline
\textbf{Methods} & \textbf{Type} & \textbf{Concepts} & \textbf{Hyperparameters} \\ \hline
OpenFace & Pre-trained   model & \begin{tabular}[t]{@{}l@{}}Gaze, \\ Vocalization, \\ Activity-level/arousal\end{tabular} & \begin{tabular}[t]{@{}l@{}}Default using \\ FeatureExtraction\end{tabular} \\ \\
FER & Pre-trained   model & \begin{tabular}[t]{@{}l@{}}Positive affect, \\ Negative emotionality\end{tabular} & Default with mtcnn=True \\ \\
BackgroundSubtractorMOG & Algorithm & \begin{tabular}[t]{@{}l@{}}Activity-level/arousal, \\ Anxiety, \\ Attention\end{tabular} & \begin{tabular}[t]{@{}l@{}}cv2.threshold with\\ thresh = 1, maxval=255\\ and cv2.THRESH\_BINARY\end{tabular} \\ \\
YOLOv5 & Pre-trained   model & \begin{tabular}[t]{@{}l@{}}Activity-level/arousal, \\ Anxiety, \\ Attention\end{tabular} & Default with yolov5n \\ \\
K-means & Algorithm & \begin{tabular}[t]{@{}l@{}}Activity-level/arousal, \\ Anxiety, \\ Attention\end{tabular} & \begin{tabular}[t]{@{}l@{}}k maximum \\ number of detected \\ persons in the video\end{tabular} \\ \hline
\end{tabular}
}
\end{table*}

\textbf{Model descriptions}

We used OpenFace \cite{Baltrusaitis2018} for gaze tracking and action units (AUs) extraction due to its open source nature and performance. For gaze tracking, we used the x and y eye gaze angles in world coordinates (i.e. \verb|gaze_angle_x|, \verb|gaze_angle_y| where the angles are in radians relative to camera position). For the facial expression recognition (FER), we used the Python package fer, that is based on~\cite{zhang2016, arriaga2017}. The FER model predicts the classes: $angry$, $disgust$, $fear$, $happy$, $neutral$, $sad$, and $surprise$ with individual scores that sum up to 1. Furthermore, the scores were multiplied with 100 (as a percentage) and averaged over the frames per each video.
For activity, we detected the body posture with focus on the upper-body of the youth with the help of YOLOv5~\cite{Redmon2016CVPR} and the anchor box coordinates for the $person$ class. We used k-means \cite{Lloyd1982} to group the coordinates of the anchor boxes from YOLOv5 since the interviewer was sometimes included in the video frames. We manually selected the group corresponding to the youth. For the motion heatmap (adaptation of \cite{heatmap}, \cite{kajabad2019people}), we used the OpenCV tools BackgroundSubtractorMOG \cite{Godbehere2012} and simple thresholding. \\
Details regarding the used methods are presented in Table \ref{table:hyperparameters}.
 
\textbf{Interpretable computed behavior codes}

We define and compute the ML based scores of the concepts as presented below. The scores have been computed for all the 74 videos.

\textbf{Gaze:}\\
The camera position changes across videos and the interviewer position is not visible in all videos, which makes gaze estimation difficult \cite{Tran2020}. Due to this hindrance, small video clips that reflect gaze were manually extracted from the entire depression chapter for each participant. We defined reflecting gaze as looking into the eyes of the interviewer. These video clips were not extracted by the mental health experts to avoid bias. A rectangle was computed from all the minimum and maximum values of the \verb|gaze_angle_x| and \verb|gaze_angle_y| over each gaze extracted video clip. The final gaze score per video was computed as a percentage of the number of gaze points present in the gaze rectangle divided by the total number of points (frames).

\textbf{Vocalization:}\\
For the vocalization, we used AUs related to the mouth \cite{Baltrusaitis2018} and defined presence as a minimum intensity of 1. We defined no vocalization (0\%) as presence of the AUs 10, 12, 14, 15, 17, 20, 23, in which the mouth is closed.
%(AU28 was not included as the intensity is missing). 
Medium vocalization (50\%) was defined as presence of AU25 (lips part) while high vocalization (100\%) was defined as presence of AU26 (jaw drop). The vocalization per video was computed as a weighted arithmetic mean as follows:
$$ \mathrm{Vocalization} = \frac{0\% * n_l + 50\% * n_m + 100\% * n_h}{N_p} $$
where, $n_l,n_m,n_h$ are the corresponding numbers of frames where low, medium, and high vocalization are present.\\
$N_p$ is the total frames per video where presence is detected ($N_p = n_l + n_m + n_h$).

\textbf{Positive affect:}\\ 
The \textit{happy} class score from the FER model was used to describe positive affect.

\textbf{Negative emotionality:}\\ The \textit{sad} and \textit{angry} class scores from the FER model were used to describe negative emotionality. Since the two classes are computed using the FER model, the values cannot be 100\% for both classes at the same time. Thus, we decided to report the maximum score of the two classes per video.

\textbf{Activity-level/arousal:}\\
We define activity-level/arousal as a composite score as follows: 
$$\mathrm{Activity\mbox{-}level/arousal} = \frac{\mathrm{Activity} + \mathrm{Vocalization} + c_{\mathrm{max}}}{3}$$
where,\\
$c_{\mathrm{max}} = \mathrm{max}(happy, angry, surprise, 100\%- neutral)$% 

\textit{\textbf{Activity.}}  We used a background subtraction method to get a foreground mask per frame. We used a simple thresholding and we chose a threshold of 1 to be conservative. We accumulate the masks to obtain the motion heatmap such that each motion heatmap has pixel values of 0 - 255. We converted these values to percentages, 0\% (0, low activity) and 100\% (255, high activity).
The activity score is the averaged value over all the percentages in the heatmap.

\textbf{Anxiety:}\\
For quantifying the anxiety, a composite score using \textit{fear} and \textit{disgust} class scores from the FER model and activity was used. We chose the highest percentage between \textit{fear} and \textit{disgust}. The score for anxiety per video is computed as follows:
$$ \mathrm{Anxiety} = \frac{\mathrm{Activity} + \max(fear,disgust)}{2} $$

\textbf{Attention:}\\
We defined attention as a percentage composite score to match the CIB definition. 
$$ \mathrm{Attention} =\frac{(100\% - \mathrm{Activity}) + (100\% - \mathrm{Anxiety})}{2} $$

\subsection{Evaluation}
As an evaluation measure, we used the percent agreement \cite{mchugh2012interrater}. We used this measure as it is a standard way to quantify interrater reliability in the context of CIB. $$ \mathrm{Percent\,agreement\,(\%)} = \frac{\mathrm{number\, agreements}}{\mathrm{total\,number\,items}} \times 100$$
The $\mathrm{total\,number\,items}$ can be either the total number of videos (percent agreement per CIB item) or the total number of CIB items (percent agreement per video). In our analysis, we only report the average of the percent agreement for the videos. Agreements are defined as CIB scores with a difference less than or equal to 1.\\
The scores computed by the ML methods are percentages and are further transformed into corresponding scores from 1 to 5 with half-points. This transformation was performed to match the human raters' scores for a fair comparison.\\
To be certified in the CIB, human raters must attain agreement of 85\% with an expert across all items in the manual and at least 13 videos with a length of 3 minutes.  

\section{Results}
\begin{figure}[h!]
 \centering
  \includegraphics[width=8.2cm]{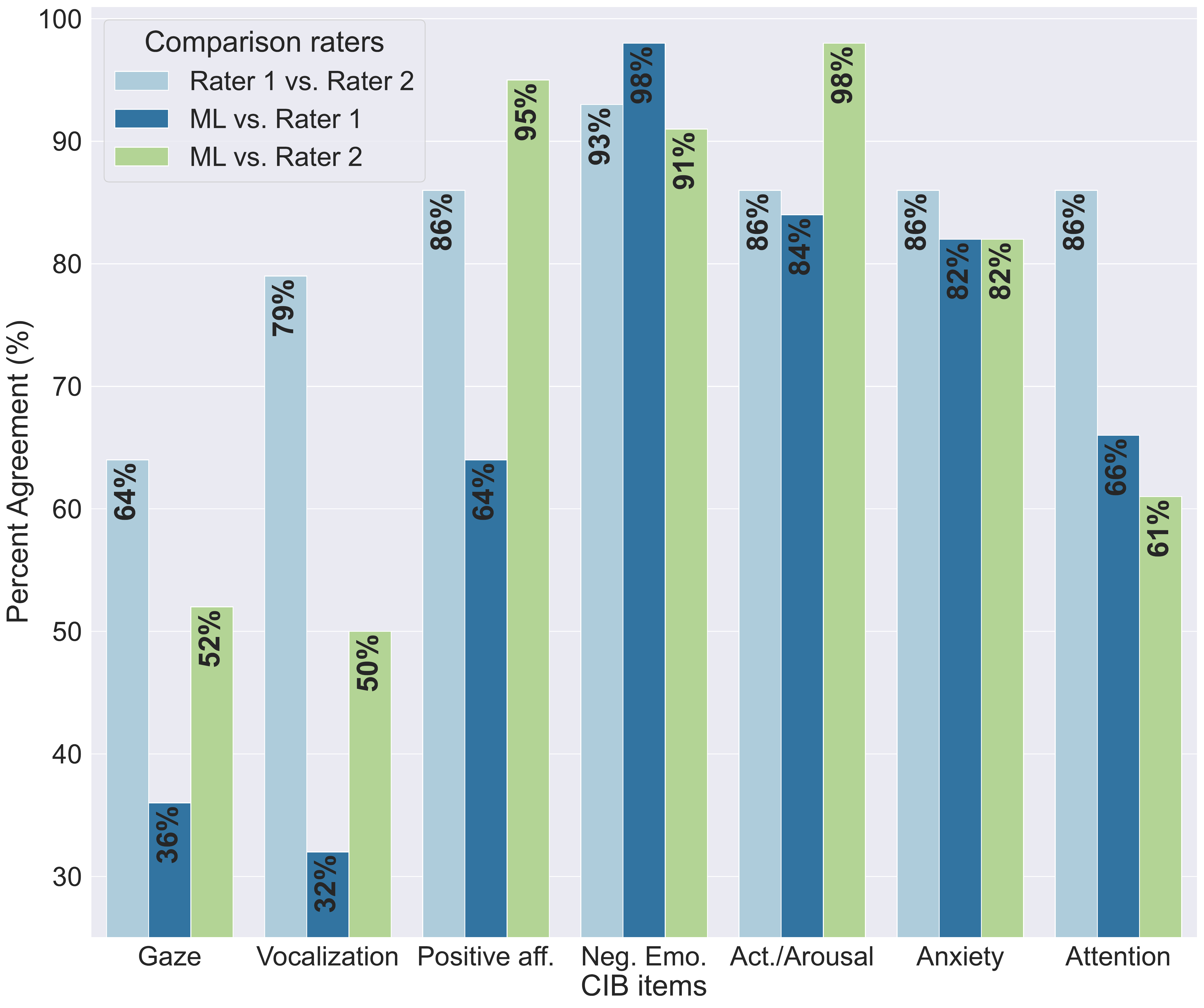}
  \vspace*{-3mm}
  \caption{Percent agreement for human Raters 1 and 2 vs. each other and vs. the machine learning approach (ML).\\ Abbreviations: positive affect (Positive aff.), negative emotionality (Neg. Emo.) and activity-level/arousal (Act./Arousal).}
  \label{fig:percentage}
  \end{figure}

Human raters had strong agreement for most codes (79-93\%), and the lowest agreement (64\%) was obtained for gaze, see Figure \ref{fig:percentage}. ML performed well on negative emotions, activity-level/arousal and anxiety where the agreement between ML and raters was similar to that between raters. ML performed reasonably on attention and positive affect, but showed a drop in agreement for these two concepts compared to that between raters. For gaze and vocalization, the performance of ML was poor with agreements between ML and raters ranging from 32\% to 52\%.
 The human raters in this study achieved 83\% agreement across all 7 CIB items, see Table \ref{table:percentage}. In contrast, the ML achieved 17\% and 7\% less agreement with the two raters, respectively. Human raters did not reach the 85\% agreement required for CIB certification for videos in this study. The CIB manual recommends rating a minimum of 3 minutes of interaction for global codes \cite{feld1998cib}. In this study, only 30 seconds were coded. 
\begin{table}[h!]
\caption{Average percent agreement over the videos}
\label{table:percentage}
\resizebox{0.9\columnwidth}{!}{
\begin{tabular}{lc}
\hline
\textbf{Comparison} & \textbf{Percentage agreement} \\ \hline
Rater 1 vs. Rater 2 & 83\%  \\
ML vs. Rater 1     & 66\%  \\
ML vs. Rater 2     & 76\%  \\ \hline
\end{tabular}
}
\end{table}
If we remove the concepts for which the ML did not show good performance with the raters, the ML achieves agreements across the remaining CIB items (positive affect, negative emotionality, activity-level/arousal, anxiety and attention), which is comparable to the agreement between the two raters (Table \ref{table:percentageDropped}).
%\vspace{-0.4cm}
\begin{table}[h!]
\caption{Average percentage agreement over the videos.\\ Dropped CIB items : gaze and vocalization}
\label{table:percentageDropped}
\resizebox{0.85\columnwidth}{!}{
\begin{tabular}{lc}
\hline
\textbf{Comparison}  & \textbf{Percent agreement} \\ \hline
Rater 1 vs. Rater 2 & 87\%  \\
ML vs. Rater 1     & 79\%  \\
ML vs. Rater 2     & 85\%  \\ \hline
\end{tabular}
}
\end{table}
\section{Discussion}
In this work, we achieved interpretability by designing models for individual behavior codes (concepts). These concepts enable evaluation of agreement with expert human raters. Building the framework on familiar concepts, gives behavioral researchers or therapists feedback in an understandable and actionable manner. For example, providing therapists feedback on the mental state of patients could inform diagnosis, gauging clinical severity or therapeutic alliance. Overall, the computationally-derived CIB scores performed well, though lower agreement was found on individual items as expected. The lowest agreement scores were obtained for gaze, vocalization, and attention. Human-machine agreement for activity-level/arousal was high despite these depending on vocalization. Thus, using lip and jaw position AUs or mouth movement increased agreement for activity-level/arousal, but these AUs did not capture the concept of vocalization. As human raters also use information from the content and sound characteristics of speech to rate behavior, a multimodal approach incorporating speech signals would likely improve machine-human agreement. To obtain better measures of gaze, uniform placement of video camera, interviewer, and subject is recommended. Low video quality is a limitation, and the low agreement between the human raters for gaze must be solved before attempting to improve the machine learning measures. Future work also includes examining interpretability of the individual ML models. 

\section{Acknowledgements}
This work is funded by the Novo Nordisk Foundation (grant number: NNF19OC0056795). The authors are thanking research assistant Anders Buch Thuesen for his contribution while temporarily supporting the project.

%\section{Conclusion}

\balance
\bibliography{example_paper}
\bibliographystyle{icml2022}

% https://tex.stackexchange.com/questions/20138/how-to-show-only-the-bibliography-which-is-cited
%\nocite{*}

\end{document}